\documentclass[%
 reprint,
superscriptaddress,
amsmath,amssymb,
aps,
prmaterials,
]{revtex4-2}

\usepackage{graphicx}% Include figure files
\usepackage{dcolumn}% Align table columns on decimal point
\usepackage{bm}% bold math
\usepackage{booktabs} % For better table lines and spacing
\usepackage{multirow} % Required if you want to use the multirow
\usepackage[table]{xcolor} % Required for coloring cells
\usepackage[colorlinks = true,
            linkcolor = blue,
            urlcolor  = blue,
            citecolor = blue,
            anchorcolor = blue]{hyperref}
\begin{document}

\preprint{APS/123-QED}
\title{Enhancing material property prediction with ensemble deep graph convolutional networks}% Force line breaks with \\
%\thanks{A footnote to the article title}%

% \author{Chowdhury Mohammad Abid Rahman}
%  \altaffiliation[Also at ]{Physics Department, XYZ University.}%Lines break automatically or can be forced with \\
% \author{Ghadendra Bhandari}%
% \author{Nasser M Nasrabadi}%
% \author{Aldo H. Romero}
% \author{Prashnna K. Gyawali}
%  \email{Second.Author@institution.edu}
% \affiliation{%
%  Authors' institution and/or address\\
%  This line break forced with \textbackslash\textbackslash
% }%

% \collaboration{MUSO Collaboration}%\noaffiliation

% \author{Charlie Author}
%  \homepage{http://www.Second.institution.edu/~Charlie.Author}
% \affiliation{
%  Second institution and/or address\\
%  This line break forced% with \\
% }%
% \affiliation{
%  Third institution, the second for Charlie Author
% }%
% \author{Delta Author}
% \affiliation{%
%  Authors' institution and/or address\\
%  This line break forced with \textbackslash\textbackslash
% }%

% \collaboration{CLEO Collaboration}%\noaffiliation
\author{Chowdhury Mohammad Abid Rahman}
\thanks{Corresponding author: cr00071@mix.wvu.edu}%
\affiliation{Lane Department of Computer Science and Electrical Engineering, West Virginia University}

\author{Ghadendra Bhandari}
\affiliation{Department of Physics and Astronomy, West Virginia University}

\author{Nasser M Nasrabadi}
\affiliation{Lane Department of Computer Science and Electrical Engineering, West Virginia University}

\author{Aldo H. Romero}
\affiliation{Department of Physics and Astronomy, West Virginia University}

\author{Prashnna K. Gyawali}
\affiliation{Lane Department of Computer Science and Electrical Engineering, West Virginia University}

\date{\today}% It is always \today, today,
             %  but any date may be explicitly specified

\begin{abstract}
Machine learning (ML) models have emerged as powerful tools for accelerating materials discovery and design by enabling accurate predictions of properties from compositional and structural data. These capabilities are vital for developing advanced technologies across fields such as energy, electronics, and biomedicine, potentially reducing the time and resources needed for new material exploration and promoting rapid innovation cycles.
Recent efforts have focused on employing advanced ML algorithms, including deep learning - based graph neural network, for property prediction.
Additionally, ensemble models have proven to enhance the generalizability and robustness of ML and DL.
However, the use of such ensemble strategies in deep graph networks for material property prediction remains underexplored. Our research provides an in-depth evaluation of ensemble strategies in deep learning - based graph neural network, specifically targeting material property prediction tasks. By testing the Crystal Graph Convolutional Neural Network (CGCNN) and its multitask version, MT-CGCNN, we demonstrated that ensemble techniques, especially prediction averaging, substantially improve precision beyond traditional metrics for key properties like formation energy per atom ($\Delta E^{f}$), band gap ($E_{g}$) and density ($\rho$) in 33,990 stable inorganic materials. These findings support the broader application of ensemble methods to enhance predictive accuracy in the field.

% \begin{description}
% \item[Usage]
% Secondary publications and information retrieval purposes.
% \item[Structure]
% You may use the \texttt{description} environment to structure your abstract;
% use the optional argument of the \verb+\item+ command to give the category of each item. 
% \end{description}
\end{abstract}

%\keywords{Suggested keywords}%Use showkeys class option if keyword
                              %display desired
\maketitle

%\tableofcontents

% \section{\label{sec:level1}First-level heading:\protect\\ The line
% break was forced \lowercase{via} \textbackslash\textbackslash}
\section{\label{sec:level1}Introduction}

Predicting material crystal properties involves forecasting the chemical and physical traits of crystalline materials based on their molecular and atomic structures. This task is vital for fields like electronics, medicine, aeronautics, and energy storage management~\cite{shen2022high,shahzad2024accelerating,wei2019machine,liu2017materials}. Accurately predicting these properties from compositional and structural data is instrumental in the discovery of new materials for advanced technologies. Using computational methods and data-driven strategies, researchers can efficiently explore and optimize material designs, avoiding the slow and costly process of experimental trial-and-error~\cite{kauwe2020extracting}.
Understanding how a material's crystalline structure impacts its properties requires a blend of computational and experimental investigations. Although density functional theory (DFT) \cite{kohn1965self,hohenberg1964inhomogeneous} is a well-established method, it is often perceived as computationally complex and time-intensive. Moreover, the quest for materials with specific properties within an extensive material search space poses challenges, and delays progress in swiftly evolving domains like medical science, aeronautics, and energy engineering, where accuracy and speed are paramount.

% Machine learning-based models have emerged as a promising solution to tackle this obstacle. These models can rival the accuracy of DFT calculations while catering to the need for rapid material discovery, facilitated by the burgeoning material databases~\cite{jain2013commentary,choudhary2020joint,kirklin2015open}. 
Machine learning (ML)-based models have emerged as a promising solution to this challenge. These models can match the accuracy of DFT calculations and support rapid material discovery, thanks to the growing material databases~\cite{jain2013commentary,choudhary2020joint,kirklin2015open}. By harnessing the capabilities of ML algorithms and the growing abundance of data in material repositories, these models can effectively navigate the vast material landscape and pinpoint promising candidates with desired properties. 
%This approach accelerates the trajectory of material exploration and advancement.
This data-driven approach for predicting material crystal properties has shown promise and gained significant attention for its accuracy and unparalleled speed \cite{choudhary2021atomistic}.%\PG{(citation)}.

However, 
%the intricate nature of crystal structures and the limitations inherent in computational models may impede machine learning (ML) approaches from fully capturing their complexity.
the intricate atomic arrangements and the intrinsic correlations between structure and properties present formidable challenges for ML models to encode pertinent structural information accurately, mainly because of the number of atoms involved and the internal degrees of freedom within the crystal structures.
% To overcome these hurdles, developing advanced ML models capable of navigating the intricacies of crystal structures while integrating domain expertise and physical principles within modest database sizes is imperative. This endeavor may entail devising innovative machine-learning frameworks, incorporating physical constraints and symmetry considerations, or exploring hybrid methodologies that consolidate machine-learning strategies with first-principles computations or human insight.
% The availability of high-quality and diverse crystal structure datasets is a crucial factor in training resilient and adaptable ML models. Therefore, sustained efforts in curating and expanding crystal structure repositories, coupled with improvements in data quality and representation, can significantly enhance the efficacy and dependability of ML models in this field.
 %The vast array of atoms in three dimensions, diverse structural prototypes, chemical compositions
 %(as the number of species increases, there is an exponential growth in the number of crystal systems)
% , and impurities, vacancies, and other crystal defects complicate the task. 
Thus, effective representation %necessitates 
capturing spatial relationships and periodic boundary conditions within a crystal lattice becomes challenging. Furthermore, a model adept with one crystal structure might falter with another, given crystals' inherent periodicity and symmetry. Also, traditional ML models often fail to incorporate the nuanced knowledge of unit cells and their repetitive nature, an essential aspect of crystallography~\cite{chen2019graph}.

Moreover, for ML models the challenge of representing crystal systems, which vary widely in size, arises because these models typically require input data in the form of fixed-length vectors. To address this, researchers have developed two main approaches. The first involves manually creating fixed-length feature vectors based on basic material properties~\cite{isayev2017universal,xue2016accelerated}, which, though effective, necessitates tailored designs for each property being predicted. The second approach uses symmetry-invariant transformations~\cite{seko2017representation} of the atomic coordinates to standardize input data, which, while solving the issue of variable crystal sizes, complicates model interpretability due to the complexity of these transformations. 
Historically, AI-driven material science has focused on creating custom descriptors for predicting material properties, utilizing expert knowledge for specific applications. However, these custom descriptors often lack versatility beyond their initial scope. 

To address these challenges in material property prediction, the recent adoption of deep learning (DL) has shown significant promise~\cite{cao2019convolutional}.
These networks are adept at learning data distributions as embeddings, serving as effective feature descriptors for predicting input data characteristics. 
Unlike traditional grid-like image representations, crystal structures—with their node-like atoms and bond-like edges—are ideally represented as graph-based structures. 
This has highlighted the suitability of graph neural networks (GNN) for modeling crystal structures, leveraging their natural composition and bonding structures. 
GNNs thus facilitate the automatic extraction of optimal representations from atoms and bonds by representing these materials in deep networks. %This method is particularly effective for applications such as molecular feature extraction and drug discovery, underscoring the potential of GNNs to tackle complex tasks involving molecular and crystal structures~\cite{sanyal2018mt}.

While current DL models, including advanced GNNs, have successfully integrated complex structural, geometrical, and topological features for predicting material properties, a critical aspect often overlooked is the comprehensive exploration of their training dynamics. Typically, the quest for the lowest validation loss serves as a proxy for identifying the optimal model. However, due to the highly non-convex nature of training deep neural networks, this lowest validation loss does not necessarily correspond to the true optimal point \cite{gyawali2022ensembling,garipov2018loss}. Other regions within the loss landscape might capture the relationship between material structure and properties differently.
There has been limited focus on understanding model behavior beyond the conventional point of lowest validation loss. This oversight suggests we might not fully grasp the potential and versatility of these models in capturing structure-to-property relationships.

In this research, we propose a critical yet often overlooked hypothesis: optimal model performance may not reside solely at the singular point of lowest validation loss. Instead, it could be spread across multiple valleys within the loss terrain. Our objective is to highlight these hidden, underexplored models that could offer more accurate predictions and a deeper understanding of material attributes. Furthermore, this approach may reveal models that provide a better balance between variance and bias, underscoring the necessity of examining the overall loss landscape to fully understand the adaptability and efficiency of deep learning models.

%In this research, we delve into the realm of GNN-based material prediction models, extending beyond the traditional focus on minimizing validation loss—a metric traditionally associated with achieving the best model performance on unseen test data.   
%We aim to investigate the complex and varied loss landscapes within which our models operate, landscapes as rich and diverse as the materials we seek to understand. 
Thus, we investigate various regions within the loss landscape where models still perform satisfactorily. This exploration has yielded insights into models that are robust and generalize well to new data. We propose combining these different models to create a unified ensemble model.
First, we analyze the property prediction performance of a prominent GNN-based material prediction model, CGCNN~\cite{xie2018crystal}, and its multi-task variant, MT-CGCNN~\cite{sanyal2018mt}, focusing on three widely studied properties: formation energy per atom, bandgap, and density. We then examine the impact of ensemble techniques on model performance, specifically the model average ensemble and prediction average ensemble methods, to support our hypothesis of identifying influential models within adjacent areas of the loss landscape.
Finally, we conduct an extensive evaluation to understand the ensemble's effect on prediction performance across the spectrum of properties. Overall, our contributions include:
\begin{enumerate}
    \item We introduce ensemble techniques to improve the material property prediction capabilities of prominent GNN-based methods, including CGCNN~\cite{xie2018crystal}, and MT-CGCNN~\cite{sanyal2018mt}.
    \item We conduct comprehensive experiments on three widely studied properties: formation energy per atom, bandgap, and density. 
    \item We assess the impact of ensemble models across a wide spectrum of material properties, highlighting the effectiveness of ensemble-based approaches in extreme test conditions. 
\end{enumerate}

\section{Related Work}
Not having regular grid-like structures such as images or 1-D signals, CNN could not be the automatic choice for studying material structure from DL point of view. Rather its irregular structural shape made it a suitable candidate for graph representation and graph neural network (GNN), where atoms are considered nodes and atomic bonds edges. Therefore, convolving on the graph structure which is converted from the actual atomic structure of the material was a prominent choice for the researchers. Crystal Graph Convolution Neural Network (CGCNN) \cite{xie2018crystal} and SchNet\cite{schutt2018schnet} first proposed this graph representation and utilized raw features like atom type and atomic bond distance to predict material properties comparable with DFT computed values. 
Although these models performed very well in terms of predicting material's properties they also exhibit inevitable and notable challenges due to their reliance on distance-based message-passing mechanisms~\cite{gong2023examining}. 
Firstly, neglecting many-body and directional information can overlook critical aspects important for understanding material properties. Secondly, the reliance on nearest neighbors to define graph edges could misrepresent key interactions due to the ambiguity of chemical bonding. Lastly, the models are limited by their receptive field, compounded by issues like over-smoothing and over-squashing, which restrict their ability to account for the long-range or global influence of structures on properties.

The more recent DL models for material property prediction such as Atomistic Line Graph Neural
Network (ALIGNN)~\cite{choudhary2021atomistic}, iCGCNN~\cite{park2020developing},MatErials Graph Network
(MEGNet)~\cite{chen2019graph}, Orbital Graph Convolution Neural Network (OGCNN)~\cite{karamad2020orbital}, DimeNet~\cite{gasteiger2020directional}, GemNet~\cite{gasteiger2021gemnet} and Geometric-information-enhanced Crystal Graph Neural Network (Geo-CGNN)~\cite{cheng2021geometric}  thus tried to incorporate more geometrical information like bond angle, orbital interaction, body order information, directional information, distance vector to outperform previously proposed distance-based models. Some models proposed attention mechanisms and self supervised learning (SSL) as well to choose the relative importance of features in predicting material properties like Matformer~\cite{yan2022periodic}, Equiformer~\cite{liao2022equiformer}, GATGNN~\cite{louis2020graph}, Crystal-Twins~\cite{magar2022crystal} and DSSL~\cite{fu2024physics}.
%\textcolor{red}{Introducing new material features, incorporating many-body or directional information, and utilizing more complex deep learning architectures have significantly improved prediction accuracy. However, these advancements come at the cost of increased computational and network complexity.}

It is a well-known fact that Deep learning models for complicated tasks navigate a high-dimensional space to minimize a function that quantifies the `loss' or error between the actual data and the expected results. We refer to this optimization landscape as the `loss landscape'~\cite{li2018visualizing,goodfellow2016deep,garipov2018loss}. It is frequently represented visually as a surface or landscape with hills and valleys. 
%Deep learning has proven successful in many application fields, but little is known about the loss surfaces. Finding the point in this landscape that corresponds to the minimal loss, or the point at which the model's predictions are as near to the actual results as feasible, is the aim of training a deep learning model. To achieve and identify the optimal model performance, there are a few subtleties introduced by the training process and the character of these landscapes. 
The loss landscapes of 
%deep learning models, particularly 
deep neural networks are extremely intricate and non-convex. This indicates that while the model may converge to numerous local minima (valleys), not all of them will result in the best solution (global minimum). 
%Models are tested on a validation set at regular intervals throughout training to track how well they perform with unknown data. 
The training epoch in which the model achieves the lowest validation loss is referred to as the `best-validated epoch'. Nevertheless, the model at this epoch may not truly represent the best generalizable model because of the complexity of the loss landscape and elements like overfitting and identical loss with differences in functional space~\cite{draxler2018essentially,fort2019large,fort2019deep}. Though they may have a little larger or the same loss, models from other epochs may perform better on unknown data or have superior generalization ability for having differences in loss dynamics. Therefore, our focus in this work is not on the GNN models or associating features with them to strengthen material property prediction but on creating a generalized framework of ensemble models based on the cross-validation loss trajectory that might yield better generalization of the prediction task with improved accuracy of prediction. 

The comprehensive approach of any ensemble technique is to accumulate a set of models 
%($M_{1}, M_{2}, M_{3}.. M_{n}$) 
using a function. It has widely been used to strengthen the performance of ML and DL models. There have been a number of approaches that can be followed to ensemble the power of various models in DL.   
%$G$. If the size of the dataset is $k$ and the dimension of the feature is m then, the dataset can be represented as, $D= {(x_{l},y_{l})}, 1\leq l \leq k,x_{p}\in \mathbb{R}^{m}$. Applying the ensemble function to predict the output of the dataset for input $x_{l}$ can be written as \eqref{eq_ensemble} 
% \begin{equation}
%     y_{l} = \Theta (x_l) = G(M_{1}, M_{2}, \ldots, M_{n})
% \label{eq_ensemble}
% \end{equation}
For example, diversity in base classifiers for ensembles is achieved through two distinct approaches, depending on whether the ensemble is composed of models of the same type (homogeneous) or different types (heterogeneous)~\cite{ganaie2022ensemble}. Different data fusion methods like max-voting, average voting, weighted average voting, and meta-learning have been proposed~\cite{mohammed2023comprehensive}. Also, four types of ensemble techniques are normally adopted in literature such as bagging, boosting, sorting, and decision fusion \cite{ganaie2022ensemble}. 
%Based on the ensemble approach, deep learning can be carried out using diverse strategies. Such as utilizing multiple basic models on the same dataset, exploring different basic model architectures on identical data, or applying a broad spectrum of basic models across various data samples. Another approach involves varying the structures within a single fundamental model to analyze a wide array of data samples, thereby enhancing the robustness and accuracy of predictive outcomes~\cite{mohammed2023comprehensive}.
%Existing literature already claims the effectiveness of ensemble learning in augmenting the accuracy of DL models~\cite{haralabopoulos2020ensemble,tasci2021voting}. 
Different ensemble techniques are already being used to improve the overall performance of the single-base models in various arenas of deep learning. Although the ensemble is a highly studied topic, it is mostly used in fields like speech recognition~\cite{li2017semi}, health-care~\cite{tanveer2021classification}, natural language processing~\cite{elnagar2020arabic}, and computer vision~\cite{roshan2024deep}. Compared to these works, our work is the first to consider the ensemble strategies for material property prediction tasks, and our ensemble framework is simple and considers the ensembling by aggregating models across different training stages.    
%The performance of any ensemble approach can typically be distilled into three key factors. First is the reliance on the underlying models, categorized by their operational sequence, either in tandem or concurrently. Secondly, the integration strategy, which encompasses the selection of an appropriate technique to merge the outputs from individual classifiers, utilizing diverse weighting schemes or meta-learning strategies. Lastly, the composition of the baseline classifiers plays a role, distinguishing between ensembles that are uniform in nature and those that incorporate a variety of model types. 
\section{Methodology}
In this section, we first provide background details about the GCNN framework for material property prediction, and then introduce our ensemble framework to achieve enhanced predictions.

\subsection{Preliminary: GCNN}
Graph Convolutional Neural Networks (GCNNs) have emerged as a powerful tool in materials science, enabling researchers to analyze and predict the properties and behaviors of materials in a novel and efficient manner. Unlike traditional convolutional neural networks that process grid-like data (e.g., images), GCNNs are designed to handle graph-structured data, which is intrinsic to the representation of atomic and molecular structures in materials science. These models exploit the graph structure of materials, where nodes can represent atoms and edges can denote chemical bonds or spatial relationships. By doing so, GCNNs can capture both the local and global structural information of materials.
\begin{figure*}[!hbpt]
\centering
\label{subfig:56}\includegraphics[width=0.8\textwidth]{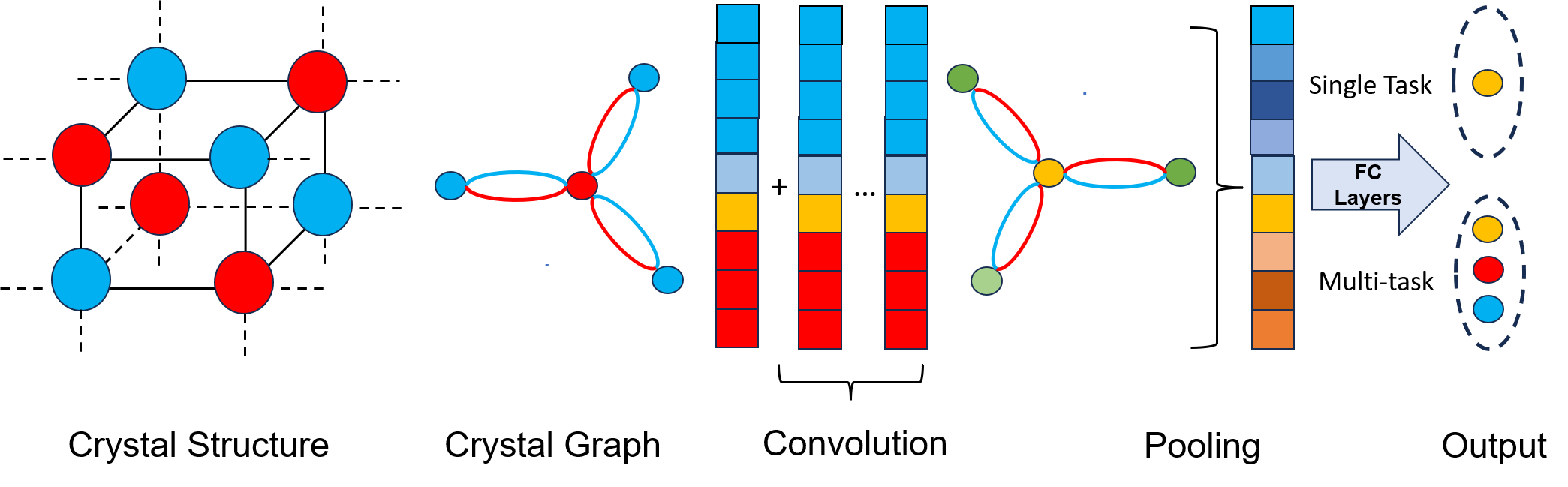}\hfill
\caption{Overview of GCNN in material property prediction tasks. Initially, the crystal structure is created from information files. Then, the crystal graph is constructed from the structure. GCN layers and pooling layers are utilized to obtain crystal embeddings, after which fully connected (FC) layers are employed to predict properties. For a single-task head, a single FC layer is used, while for the prediction of multiple properties, multiple FC layers are utilized.}
\label{model_1}
\end{figure*}

Crystal Graph Convolutional Neural Networks (CGCNN)~\cite{xie2018crystal} and SchNet~\cite{schutt2018schnet} represent the two most prominent graph neural network architectures tailored for material science applications. These models refine atom representations within a structure by considering the types and bond lengths of neighboring atoms. Subsequently, they aggregate these updated atom-level representations to form a comprehensive representation of the entire structure. In CGCNN, the crystal architecture is represented as a graph that accommodates the details of atoms and their bonds with neighbors, and a graph 
convolution network is constructed on such graph to attain the representations useful to material property prediction.
The architecture can comprise of single-task head \cite{xie2018crystal} or multi-task \cite{sanyal2018mt} head depending on the application. 
We present the overall diagram combining the single- and multi-task setup in Fig. \ref{model_1}.
In the presented network, the atom feature encoding vector can be noted as $v_{i}$ where i is an atom or node in the graph. Similarly, an edge or atomic bond among neighbors is denoted as $(i,j)_{k}$ representing $k^{th}$ bond between atoms $i$ and $j$, and its feature vector is $u_{(i,j)_{k}}$. The goal here is to update the atom feature vector $v_{i}$ by iterative convolution with neighbors and bonds~\eqref{eq_conv} and generate a comprehensive feature vector for the crystal structure by pooling~\eqref{eq_pool}.  
\begin{equation}
v^{(t+1)}_i = \text{Conv}\left( v^{(t)}_i, v^{(t)}_j, u_{(i,j)k} \right), \quad (i,j) \in \mathcal{G}.
\label{eq_conv}
\end{equation}
\begin{equation}
v_c = \text{Pool}\left( v^{(0)}_0, v^{(0)}_1, \ldots, v^{(0)}_N, \ldots, v^{(R)}_N \right)
\label{eq_pool}
\end{equation}
\begin{equation}
\min_{\mathbf{W}} J(y, f(\mathbf{C}; \mathbf{W}))
\label{eq_optim}
\end{equation}

The training procedure involves minimizing a cost function $J(y; \hat{y})
$ where $y$ is the DFT computed value and $\hat{y}$ is the prediction of the model. Therefore, CGCNN can be considered a function that tries to approximate the actual property value $y$ by mapping a crystal $C$ employing weights $W$ as shown in Eq. \eqref{eq_optim}.

\subsection{Ensemble models}
In this section, we present an ensemble framework to enhance the results obtained from GCNN networks. Our central hypothesis is that in the training of deep neural networks, selecting a single model based solely on the lowest validation error may not always guarantee the most comprehensive learning of all features within the dataset. This limitation can be attributed to the highly non-convex nature of the loss landscape that characterizes neural network optimization \cite{cooper2018loss}. In such a complex terrain, the path to minimizing loss involves numerous local minima and saddle points. This suggests that multiple models—each residing in different areas of the loss landscape—could perform similarly well on the validation set, albeit potentially capturing different aspects of the data \cite{gyawali2022ensembling}.

Toward this, 
we propose ensemble framework for achieving ensemble models from the training dynamics of 
GCNNs, that capitalizes on the temporal evolution of model parameters across training epochs. Consider the training process to span a fixed number of epochs, $T$, during which each epoch yields a candidate model characterized by unique properties and attributes reflective of its learning state at that point in time. 
Let, $f(\mathbf{x}; \Theta_t)$ represent the GCNN model at some epoch $t$. Here $x$ is the input data and $\Theta_t$ represents the model's parameters (weight and biases) at epoch $t$. 
By running the training for $T$ epochs, we generate the sequence of models $\{ f(x; \Theta_1), f(x; \Theta_2), \ldots, f(x; \Theta_T) \}$, and for each model $f(\mathbf{x}; \Theta_t)$, we compute the MSE on a given validation set as:
\begin{equation}
    \text{MSE}_t = \frac{1}{N_{\text{val}}} \sum_{i=1}^{N_{\text{val}}} \left( y_i - f(\mathbf{x}_i; \Theta_t) \right)^2
\label{eq_mse}
\end{equation}
where $N_{val}$ is the number of sample in the validation set, $y_{i}$ is the true value and $f(x_{i}; \Theta_t)$ is the predicted value for the $i^{\text{th}}$ sample in the validation set by the model at epoch $t$. 
Using the metric from Eq. \ref{eq_mse}, we sort the models based on their MSE, selecting the top $n$ models with the lowest MSE values. 
Let $\Theta^{(1)}, \Theta^{(2)}, \ldots, \Theta^{(n)}$ be the parameters of these top-$n$ models after sorting, and we present two 
different strategies for aggregating these $n$ models.
\subsubsection{Prediction-based ensemble modeling} For prediction-based ensemble modeling, we first calculate the prediction for each $\Theta^{(t)}$ model, within the top-$n$ as:
\begin{equation}
\hat{y}_{t} = f(x; \Theta_{t})
\label{eq_pred}
\end{equation}
and create prediction ensemble as:
\begin{equation}
\bar{y}_{\text{prediction-ensemble}} = \frac{1}{n} \sum_{i=1}^{n} \hat{y}_{n}
\label{eq_mae_pred_en2}
\end{equation}

\subsubsection{Model-based ensemble modeling} Here, we first aggregate top-$n$ models together by creating an ensemble model $\Theta_{avg}$:
\begin{equation}
\Theta_{\text{avg}} = \frac{1}{n} \sum_{j=1}^{n} \Theta^{(j)}
\label{eq_mod}
\end{equation}
The final prediction of the ensemble model for a new input $x$ is then given as:
\begin{equation}
\hat{y}_{\text{model-ensemble}} = f(x; \Theta_{\text{avg}})
\label{eq_pred1}
\end{equation}

\begin{figure*}[!htbp]
\centering
\label{subfig:66}\includegraphics[width=0.99\textwidth]{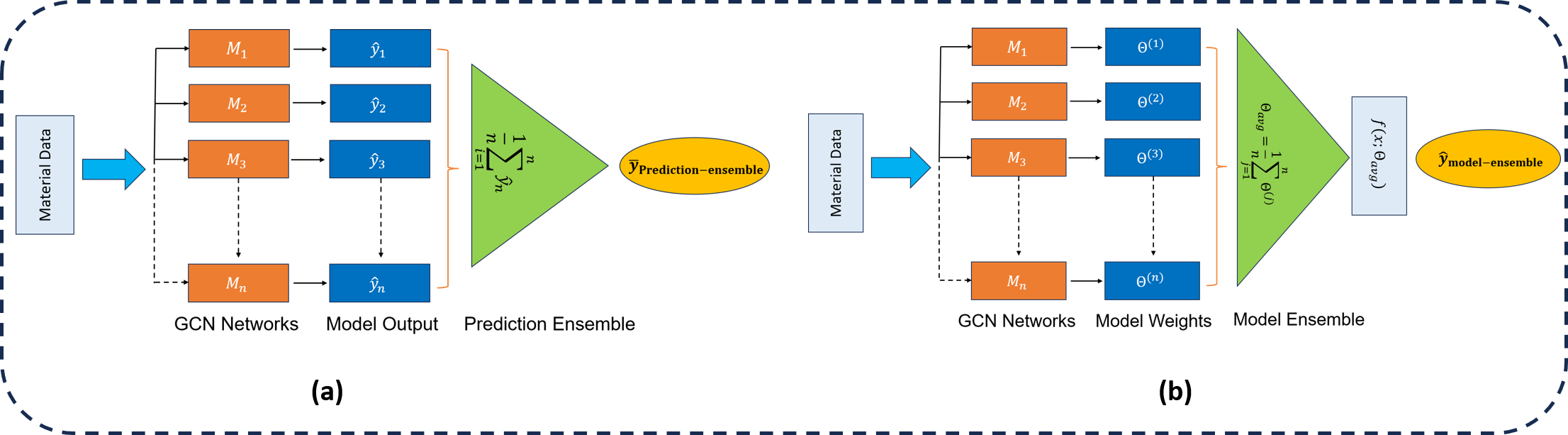}\hfill
\caption{Overview of ensemble strategies: (a) prediction averaging ensemble technique and (b) the model averaging ensemble technique.}
     \label{model_2}
\end{figure*}

We present the overall schematic of our proposed ensemble-based framework in Fig. \ref{model_2}. On the left, we illustrate the prediction ensemble, and on the right panel, the model ensemble framework is depicted.

%%%%%%%%%%%%%%%%%%%%%%%%%%%%%%%%%%%%%
\section{Experiment}
% \PG{Write a one or two line statement saying how we first talk about data and then present our experimental setup before discussing about our results.}
In this section, we discuss the data we have used for experiments, the implications of choosing the properties we worked with, the outline of the experimental setup, and the results with their interpretation, significance, and implications. 
\subsection{Data}
All the models in this study were trained on the dataset from the Materials Project \cite{jain2013commentary}. The Materials Project is a multi-institutional, multinational effort to compute the properties of all inorganic materials and provide the data and associated analysis algorithms to every materials researcher free of charge. One of the largest and most popular three-dimensional (3D) materials datasets in the materials science field, the Materials Project collection covers approximately 155,000 materials, including a wide variety of inorganic compounds. This broad coverage guarantees a representative and varied sampling of material kinds, extending the generalizability of our results. Moreover, the MP dataset has been successfully used in numerous studies to create and evaluate predictive models for a range of material properties. The trustworthiness of the dataset and the accomplishments of earlier studies employing MP data highlight its suitability for verifying our methods. For this study, we worked with 33,990 stable materials, which refers to the set of materials that, under standard conditions, have a low energy state and are hence thermodynamically favorable to exist in their current form. An indicator of stability is the `energy above hull' metric, which shows how much energy would be released by each atom in the event that the material changes into the most stable phase combination. For several reasons, it is critical for this work to concentrate on stable materials. Stable materials are more important for real-world applications because they are thermodynamically favored, meaning they occur naturally or can be created with less energy. Focusing on these materials allows the research to directly target materials that are useful in energy storage, electronics, and catalysis, among other real-world applications. To demonstrate the efficacy of the proposed ensemble framework, we focused on three distinct properties: Formation energy per atom ($\Delta E^{f}$), Density ($\rho$), and Bandgap ($E_{g}$). These three properties are important because formation energy determines the thermodynamic stability of a material, bandgap signifies whether a material is an insulator, conductor, or semiconductor, and density defines the stiffness. 
%%%%%%%%%%%%%%%%%%%%%%%%%%%%%%%%%%%
\subsection{Experimental setup}
For exploring the efficacy of our proposed ensemble model, we primarily consider CGCNN \cite{xie2018crystal} and its multi-task extension, MT-CGCNN~\cite{sanyal2018mt}, as base models, and apply our proposed ensembling to construct the ensemble framework. Furthermore, we vary the number of convolutional layers within CGCNN and MT-CGCNN to create two separate versions of the network for each category. Throughout our experiments and results, we refer to them as CGCNN$_{\text{1}}$ (number of convolutional layers, $n{\text{c}}$ = 3), CGCNN$_{\text{2}}$ ($n{\text{c}}$ = 5), MT-CGCNN$_{\text{1}}$ ($n{\text{c}}$ = 3), and MT-CGCNN$_{\text{2}}$ ($n{\text{c}}$ = 5). For the MT-CGCNN models, our multi-task objective involved predicting all three properties together using three separate heads, and optimization weights of 1.5 were used for $\Delta E^{f}$, 3 for $E_{g}$, and 1.5 for $\rho$. 
Although we present both prediction and model ensemble frameworks, we found the prediction ensemble to yield the best results. Therefore, we utilize the prediction ensemble for all our analyses and comparisons against the Best-val model. However, it is important to note that we also compare the performance between prediction and model-based ensembles to establish the efficacy of the prediction average ensemble over the latter.
% \PG{Also, point out that although we present both prediction and model ensemble framework, we found prediction ensemble to produce the best result so we consider prediction ensembe for all our analysis and comparisions against baseline. But poitn out that we also compare the performance between prediction and model based ensemble.}

Across all models, the length of the atom feature vector was set to 64, and the hidden feature vector to 128. For all experiments, MSE was used as the loss function, SGD as the optimizer, and a fixed learning rate of 0.01 was applied. We utilized an NVIDIA GeForce GTX TITAN X graphics processing unit (GPU) with 12 GB of memory for all model training and evaluation tasks. The training, validation, and test data were randomly selected from 33990 data points and were kept consistent across all experiments, employing a 70-10-20 distribution strategy, and the batch size was uniformly set at 256. This random selection ensures that our model is not biased towards any specific subset of data, thereby enhancing its ability to generalize to unseen data.  Our sampling strategy did not follow any particular distribution, ensuring that the training, validation, and test sets were representative of the overall dataset. By not constraining the sampling process to a specific distribution, we mitigate the risk of overfitting to particular patterns in the training data, thus improving the model's robustness. All models were run for 100 epochs, and up to 50 models were considered for the ensemble. To determine the best model among the baseline models, and to select several models for creating the ensemble, MSE loss on validation data was used. However, for reporting results, we used MAE on the test dataset to follow standard practice in the literature.
 \begin{table*}[!t]
\centering
    \renewcommand{\arraystretch}{1}
\setlength{\tabcolsep}{12pt}
\caption{Comparison of Best-val and Ensemble for CGCNN$_{1,2}$ (20 models)}
\label{tab:main_results_1}
\begin{tabular}{ccc cc cc cc cccc}
\toprule
\toprule
\addlinespace

    & \multicolumn{2}{c}{CGCNN$_{\text{1}}$}               & \multicolumn{2}{c}{CGCNN$_{\text{2}}$}                 
    &  \\ \cmidrule{1-6}
    & \multicolumn{1}{c}{Best-val} & Ensemble & \multicolumn{1}{c}{Best-val} & Ensemble
    & \\ \cmidrule{1-6}
Formation Energy ($\Delta E^{f}$) & \multicolumn{1}{c}{0.058}     & \textbf{0.054} (6.90\% {\color{green}$\uparrow$})     & \multicolumn{1}{c}{0.060}           &\textbf{0.055} (8.33\% {\color{green}$\uparrow$})      \\          
BG ($E_{g}$)  & \multicolumn{1}{c}{0.322}      & \textbf{0.301} (6.90\% {\color{green}$\uparrow$})              & \multicolumn{1}{c}{0.293}           & \textbf{0.278} (5.12\% {\color{green}$\uparrow$})  \\           
Density ($\rho$)   & \multicolumn{1}{c}{0.134}      & \textbf{0.128} (4.47\% {\color{green}$\uparrow$}))              & \multicolumn{1}{c}{\textbf{0.145}}           & 0.146 (0.69\% {\color{red}$\downarrow$})  \\
%\hline
%\hline
\addlinespace
\bottomrule
\bottomrule

%\hline

\end{tabular}
\end{table*}
%%%%%%%%%%%%%%%%%%%%%%%%%%%%%%%%%%%
\begin{table*}[!htbp]
\centering
    \renewcommand{\arraystretch}{1}
\setlength{\tabcolsep}{12pt}
\caption{Comparison of Best-val and Ensemble for MT-CGCNN$_{\text{1}, \text{2}}$ (40 models)}
\label{tab:main_results_2}
\begin{tabular}{ccc cc cc cc cccc}
\toprule
\toprule
\addlinespace
%\cline{1-5}
    & \multicolumn{2}{c}{MT-CGCNN$_{\text{1}}$}               & \multicolumn{2}{c}{MT-CGCNN$_{\text{2}}$}                 
    &  \\ \cmidrule{1-6}
    & \multicolumn{1}{c}{Best-val} & Ensemble & \multicolumn{1}{c}{Best-val} & Ensemble
    & \\ \cmidrule{1-6}
Formation Energy ($\Delta E^{f}$) & \multicolumn{1}{c}{0.082}     & \textbf{0.073} (11\% {\color{green}$\uparrow$})     & \multicolumn{1}{c}{0.081}           &\textbf{0.076} (6.17\% {\color{green}$\uparrow$})      \\          
BG ($E_{g}$)  & \multicolumn{1}{c}{0.316}      & \textbf{0.307} (2.85\% {\color{green}$\uparrow$})              & \multicolumn{1}{c|}{0.294}           & \textbf{0.280} (4.76\% {\color{green}$\uparrow$})  \\           
Density ($\rho$)   & \multicolumn{1}{c}{0.216}      & \textbf{0.192} (11.11\% {\color{green}$\uparrow$})              & \multicolumn{1}{c}{0.205}           & \textbf{0.190} (7.32\% {\color{green}$\uparrow$})  \\
%\hline
\addlinespace
\bottomrule
\bottomrule

\end{tabular}
\end{table*}
%%%%%%%%%%%%%%%%
\begin{figure}[!htbp]
\centering
\label{subfig:1}\includegraphics[width=0.99\columnwidth]{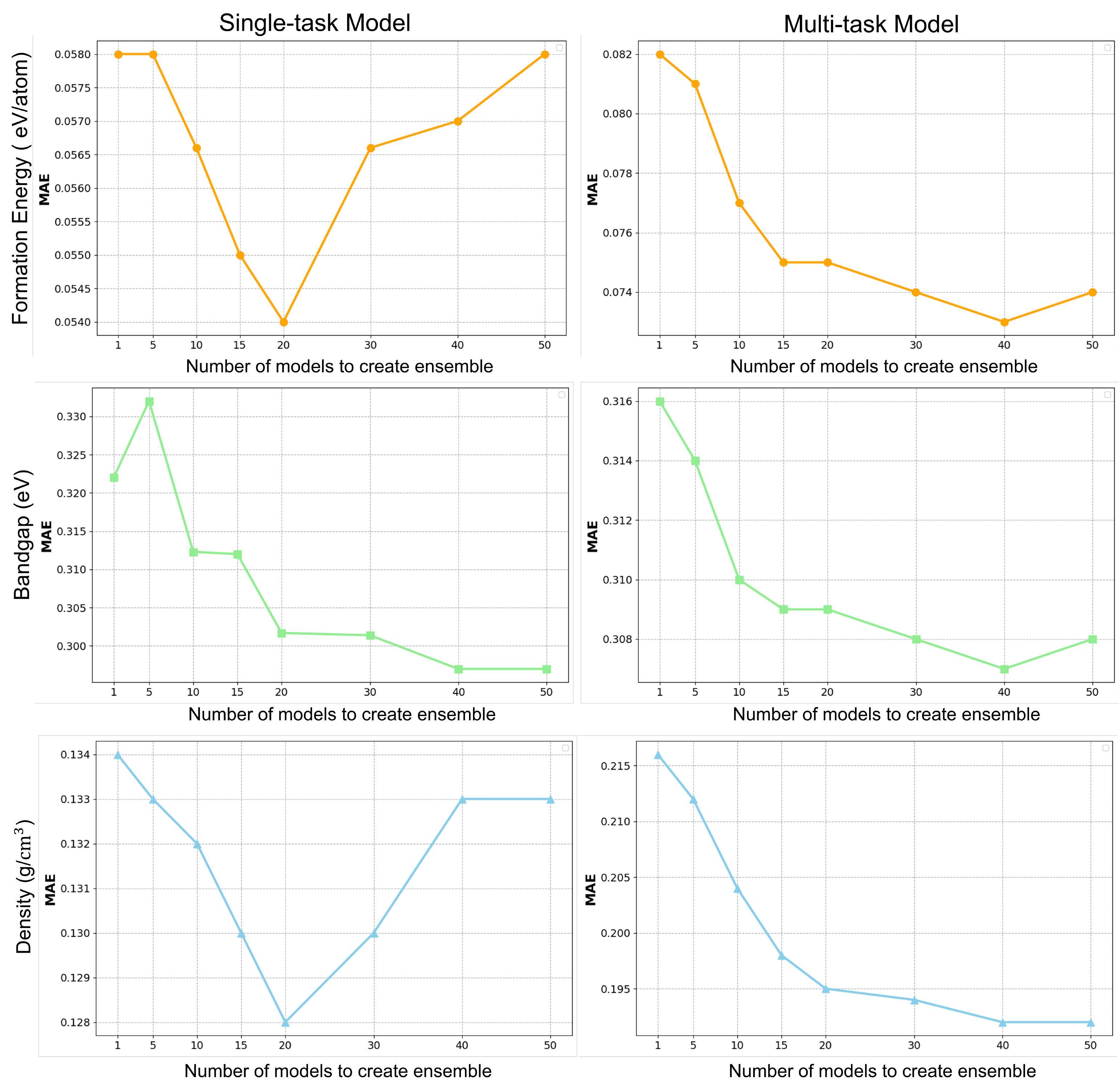}\hfill
\caption{Results for Prediction Ensemble on Single-task CGCNN and Mulit-task CGCNN. }%\PG{Also add the label for x-axis. NUmber of models for creating ensemble model. }}
     \label{ensemble_fig_1}
\end{figure}
\section{Result}

We present our main results in Table \ref{tab:main_results_1} and Table \ref{tab:main_results_2}. For results in both tables, we present outcomes from the prediction ensemble for a fixed number of epochs (20 for single-task and 40 for multi-task) as determined by validation performance. In Table \ref{tab:main_results_1}, we compare the ensemble framework (prediction ensemble) against CGCNN$_{\text{1}}$ and CGCNN$_{\text{2}}$ across all three properties. We found that, out of six different settings, our proposed ensemble framework achieved better results in five settings. Moreover, in one of the settings where the standard approach achieved a better result, the gap between our results and the baseline was the smallest (0.69\%) compared to other margins (4\% to 8\%). In Table \ref{tab:main_results_2}, we observed enhanced results by our proposed framework in all six different settings, with the improvement margin up to 11\%.
\begin{figure}[!htbp]
\centering
\label{subfig:4a}\includegraphics[width=0.98\columnwidth]{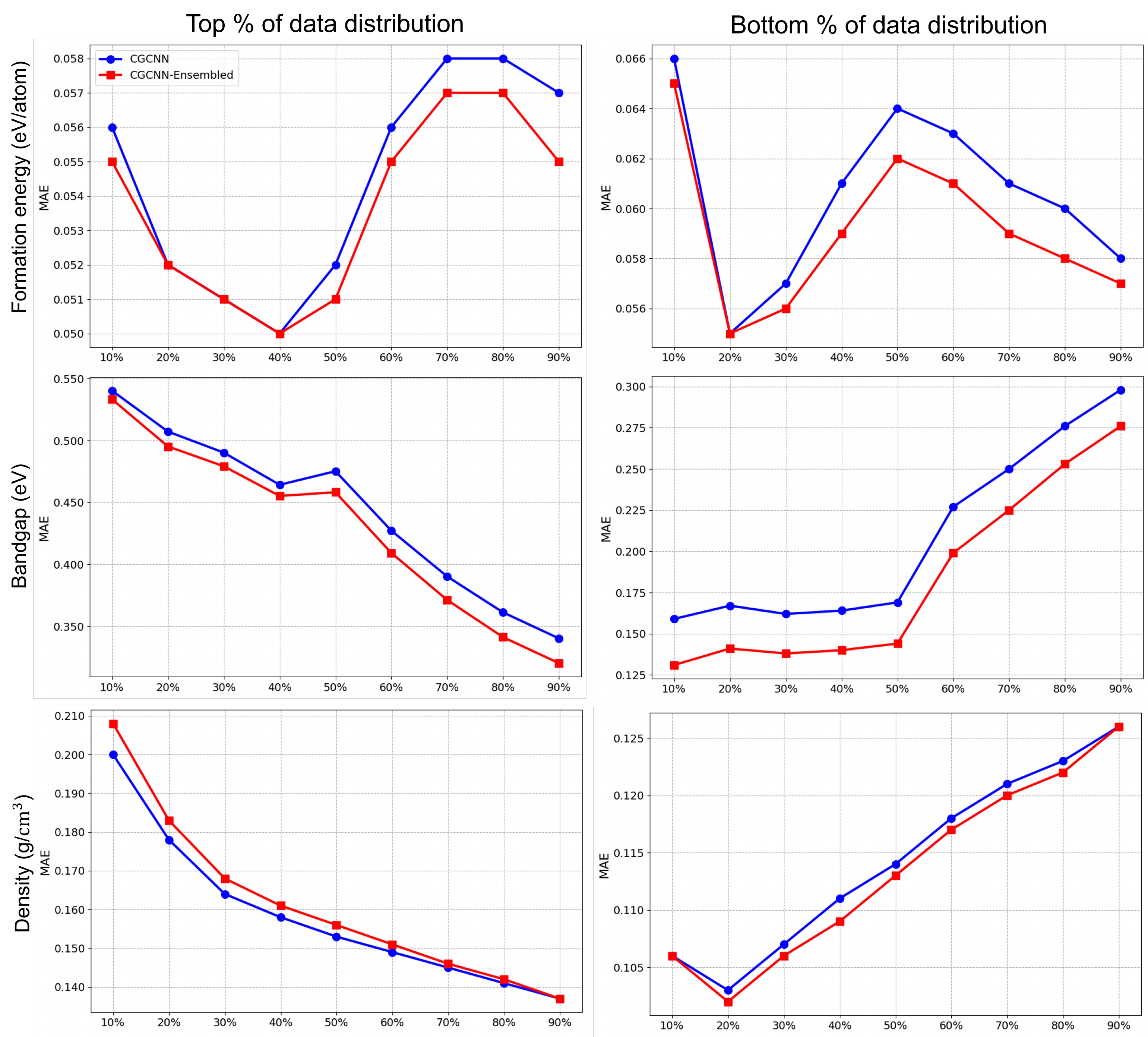}\hfill

\caption{Prediction ensemble CGCNN$_{\text{1}}$ across different groups of data distribution for three properties.} %\PG{Top \% of data distribution and Bottom \% of data distribution for header.}}
     \label{range_1}
\end{figure}
In Fig.~\ref{ensemble_fig_1}, we analyze the effect of the prediction ensemble across a different number of models. The leftmost point, representing the number of models used to create an ensemble as one, is the best validation model and represents the standard practice of validation using a single model. Compared to that single point, every result to the right represents our proposed approach of using an ensemble-based framework. As seen in both single-task (left panel) and multi-task (right panel) and across all three properties, the effect of ensembling for enhancing property prediction is quite evident. It should be noted that in some cases, such as the formation energy per atom for the single-task model, the behavior of the ensemble framework appears not to be as effective after a certain number, but we can still see that the ensemble framework is always better or similar to the best validation model. This also underscores the importance of the ensemble framework in achieving enhanced prediction results. 
\begin{figure}[!htbp]
\centering
\label{subfig:5a}\includegraphics[width=0.98\columnwidth]{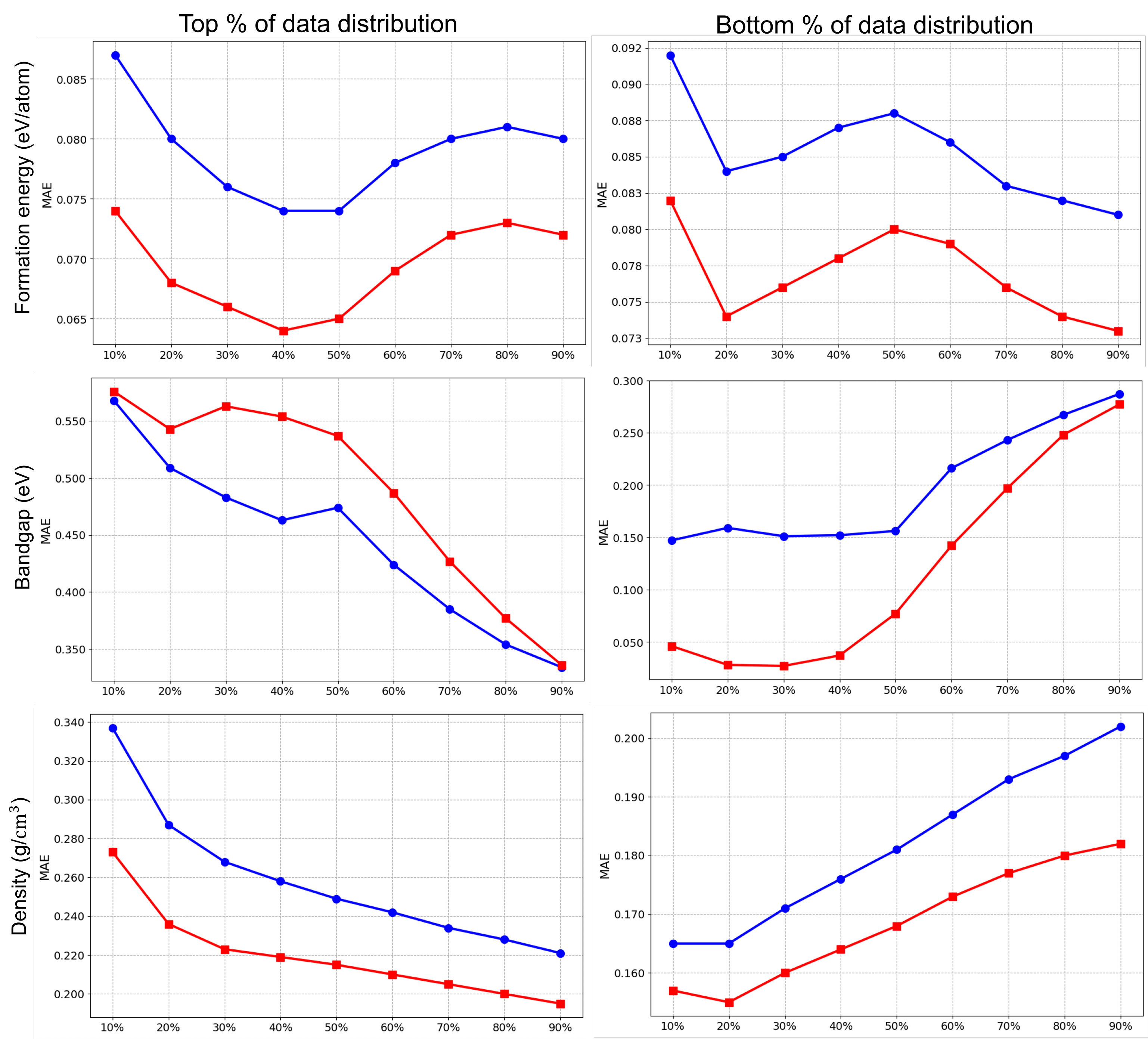}\hfill

\caption{Prediction ensemble MT-CGCNN$_{\text{1}}$ across different groups of data distribution for three properties.}
     \label{range_2}
\end{figure}

We also explore the impact of the ensemble-based framework across different regions of the material property spectrum. This analysis is crucial for a comprehensive understanding of the various properties involved in our study. For example, specific regions of the bandgap determine a material's suitability as a conductor, semiconductor, or insulator. Grasping the range of formation energy per atom is vital, as it reflects the stability and synthesizability of materials. Materials that are thermodynamically stable, and therefore more likely to occur in nature or be successfully synthesized in the laboratory, are distinguished by lower formation energies. Additionally, pinpointing the extremes within density ranges aids in assessing the durability of high-density materials or the practicality of lightweight insulators with low density.

We present the results in Fig.\ref{range_1} and Fig.\ref{range_2}, where we partitioned the test data from the 10th to the 90th percentile in both top-bottom (left panel) and bottom-top (right panel) distributions. This approach helps to identify performance differences across different regions of the property spectrum for CGCNN$_{\text{1}}$ and MT-CGCNN${_\text{1}}$ models. It is observed that the ensemble model, in all instances, aligns with the trend of the original single best model, exhibiting a reduced MAE value or improved accuracy across all percentiles of data distribution in most scenarios. Notably, significant improvement is observed in certain regions, for example, for the bottom 10\% of bandgap materials in multi-task models.

\begin{figure}[!t]
\centering
\label{subfig:2}\includegraphics[width=0.98\columnwidth]{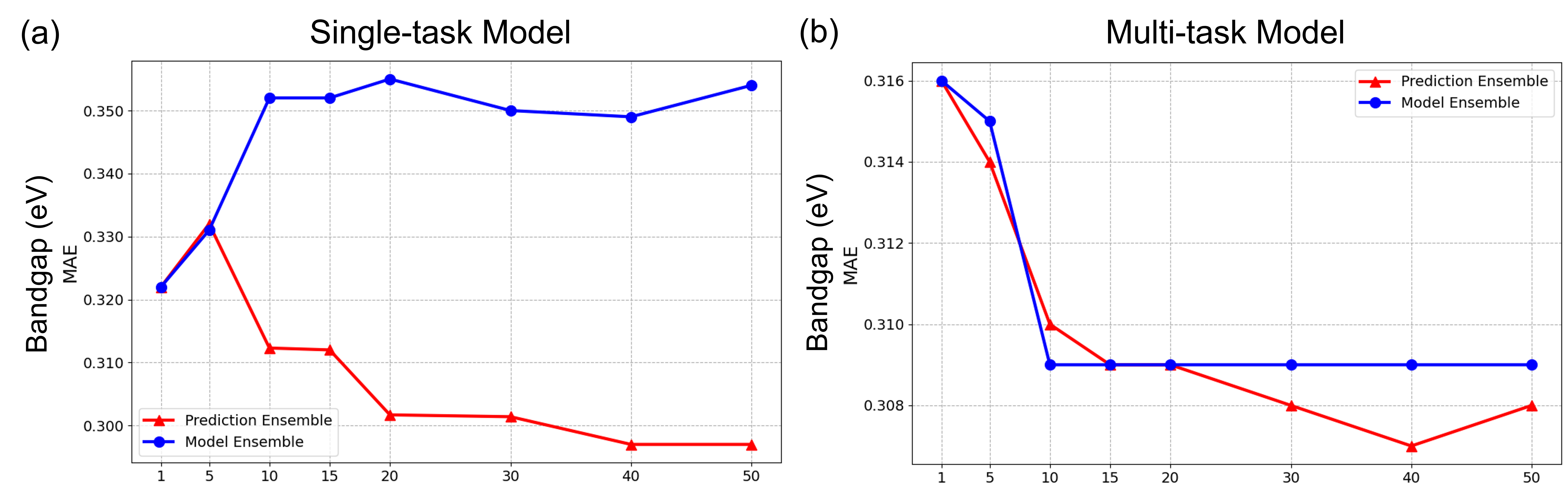}\hfill
\caption{Result comparison for Prediction Ensemble and Model average Ensemble on Single-task CGCNN$_{1}$ (a) and MT-CGCNN$_{1}$ (b) for Band gap.} %\PG{Increase the size of axes, remove the label, merge them in powerpoint!}}
     \label{ensemble_fig_2}
\end{figure}

Finally, although our experiment clearly demonstrated the benefit of a prediction ensemble over a model ensemble, in Fig.~\ref{ensemble_fig_2}, we include an analysis in this paper that compares the performance between a prediction ensemble and a model ensemble for both single-task and multi-task frameworks for band gap. As shown in the figure, although the performances of both ensemble approaches appear similar in the multi-task settings (right panel), for the single-task model, we observe that the model ensemble resulted in the worst performance, even when compared to the single best validation model.
\section{Conclusion}
In this paper, we explore the impact of an ensemble framework on the task of material prediction. Our proposed framework is both simple and effective, leveraging the loss landscape of deep neural network training without requiring computationally expensive ensemble strategies. We present two types of ensembles: prediction-based and model-based. The former involves aggregating predictions across different models to form the ensemble model, while the latter aggregates the models first, then generates a single prediction as the ensemble result. Our analysis shows that the prediction ensemble consistently outperforms the model ensemble. As a result, we conducted a comprehensive analysis across various models (single-task vs. multi-task), architectures (variations in GCNN depth), and properties, finding that the prediction ensemble almost always improves the predictive performance over the single best validation model. We also examined the efficacy of our proposed framework across different property spectra for test data distribution. Overall, our extensive analysis demonstrates the robustness of the proposed framework in generating enhanced predictive results. Future work will focus on investigating systematic approaches for calculating the number of models to select candidate models for the ensemble framework,  extending the analysis to include other GNN models, and on expanding our results to 2D datasets.

\begin{acknowledgments}
% We wish to acknowledge the support of the author community in using
% REV\TeX{}, offering suggestions and encouragement, testing new versions,
% \dots.
This research is partially supported by West Virginia Higher
Education Policy Commission’s Research Challenge Grant Program 2023 and CITeR NSF/IUCRC center (23F-02W).
\end{acknowledgments}

\bibliography{apssamp}% Produces the bibliography via BibTeX.

\end{document}